\documentclass[preprint,5p,times,twocolumn,authoryear]{elsarticle}
\usepackage{amssymb}
\usepackage{amsmath,amsfonts}
\usepackage{algorithm,tabularx} 
\usepackage{algorithmicx}
\usepackage{times}
\usepackage{array}

\usepackage[caption=false,font=normalsize,labelfont=sf,textfont=sf]{subfig}
\usepackage{textcomp}
\usepackage{varwidth}% http://ctan.org/pkg/varwidth
\usepackage{stfloats}
\usepackage{url}
\usepackage{verbatim}
\usepackage{graphicx}
\usepackage{multirow} %tabllular
\usepackage{booktabs}
\usepackage{epsfig}
\usepackage{xcolor}
\usepackage[noend]{algpseudocode}
\usepackage{amsthm}

\begin{document}

\begin{frontmatter}



\cortext[cor1]{Corresponding author.}
\title{FedMEKT: Distillation-based Embedding Knowledge Transfer for Multimodal Federated Learning}


% \author{}
\author[label1]{Huy Q. Le }
\ead{quanghuy69@khu.ac.kr}
\author[label3]{ Minh N. H. Nguyen }
\ead{nhnminh@vku.udn.vn}
\author[label1]{Chu Myaet Thwal}
\ead{chumyaet@khu.ac.kr}
\author[label2]{Yu Qiao}
\ead{qiaoyu@khu.ac.kr}
\author[label2]{Chaoning Zhang}
\ead{chaoningzhang1990@gmail.com}
\author[label1]{Choong Seon Hong \corref{cor1}}
\ead{cshong@khu.ac.kr}

% \affiliation{organization={},%Department and Organization
\affiliation[label1]{organization={Department of Computer Science and Engineering, Kyung Hee University},%Department and Organization
            city={Yongin-si},
            postcode={17104}, 
            country={Republic of Korea}}
\affiliation[label2]{organization={Department of Artificial Intelligence, Kyung Hee University},%Department and Organization
            city={Yongin-si},
            postcode={17104}, 
            country={Republic of Korea}}
\affiliation[label3]{organization={Digital Science and Technology Institute, The University of Danang—Vietnam-Korea University of Information and Communication Technology},%Department and Organization
            city={Da Nang},
            postcode={550000}, 
            country={Vietnam}}

\begin{abstract}
 Federated learning (FL) enables a decentralized machine learning paradigm for multiple clients to collaboratively train a generalized global model without sharing their private data. Most existing works have focused on designing FL systems for unimodal data, limiting their potential to exploit valuable multimodal data for future personalized applications. Moreover, the majority of FL approaches still rely on labeled data at the client side, which is often constrained by the inability of users to self-annotate their data in real-world applications. In light of these limitations, we propose a novel multimodal FL framework, namely FedMEKT, based on a semi-supervised learning approach to leverage representations from different modalities. To address the challenges of modality discrepancy and labeled data constraints in existing FL systems,  our proposed FedMEKT framework comprises local multimodal autoencoder learning, generalized multimodal autoencoder construction, and generalized classifier learning.
    Bringing this concept into the proposed framework, we develop a distillation-based multimodal embedding knowledge transfer mechanism which allows the server and clients to exchange joint multimodal embedding knowledge extracted from a multimodal proxy dataset. Specifically, our FedMEKT iteratively updates the generalized global encoders with joint multimodal embedding knowledge from participating clients through upstream and downstream multimodal embedding knowledge transfer for local learning. Through extensive experiments on three multimodal human activity recognition datasets, we demonstrate that FedMEKT not only achieves superior global encoder performance in linear evaluation but also guarantees user privacy for personal data and model parameters while demanding less communication cost than other baselines. 
\end{abstract}


% \begin{highlights}
% \end{highlights}


\end{frontmatter}


% \section{}
% \label{}
\section{Introduction}

With the rapid emergence of technologies, artificial intelligence (AI) has made significant progress in a variety of applications such as virtual assistants, e-commerce, healthcare, and recommendation systems~\citep{pawar2020explainable,lugano2017virtual,rahayu2022systematic,bawack2022artificial}, which in turn demand a massive amount of personal data from end-users. Consequently, data privacy and security become great barriers in centralized machine learning systems where the server collects personal data and trains a generic model for data-intensive applications~\citep{liu2021machine}. To tackle these challenges, federated learning (FL) has been proposed as a decentralized machine learning paradigm that aggregates model parameters from multiple clients for collaborative training without sharing their private data, thus protecting sensitive user information \citep{mcmahan2017communication,li2020federated,kairouz2021advances}. Over a decade, FL has been an engaging and demanding field of research, demonstrating promising results in various domains like smart cities~\citep{zheng2022applications,ramu2022federated,pandya2023federated} and healthcare~\citep{xu2021federated,antunes2022federated,nguyen2022federated}.

Despite many advantages in preserving privacy, existing FL methods consider a scenario where clients hold only unimodal data~\citep{mcmahan2017communication,li2020federated}, restricting the use of \textit{multimodal data} in various equipment. Recent works on deep multimodal learning demonstrate that common features from multimodal data assure more accurate and robust performance than unimodal data in many applications, such as text and image for language translation \citep{rajendran2015bridge}, audio and video for emotion recognition~\citep{liang2018computational}, and different sensor data in healthcare \citep{garcia2018mental}. As a result, multimodal data offers a broader potential for future FL applications, that utilizes multimodal client data generated from various sources and devices, including smartphones and smartwatches. In the context of multimodal FL, some prior works have shown their potential in leveraging the benefits of multimodal data for decentralized machine learning systems. The authors in~\citep{xiong2022unified} designed the co-attention layer to fuse representations from different modalities and obtain the generalized features to train personalized models. This method requires all users to have labeled data from all modalities, which means that users have to annotate the data. However, in the real world, this could be a cost hindrance for users to collect the labeled data from different modalities, such as various types of sensor data~\citep{alonso2015challenges}. Besides, \textit{label annotation} often requires a high cost and poses a significant challenge in real-world applications, particularly in healthcare, where certain sites may lack resources or incentives for extensive data labeling. One possible solution for saving the annotation cost is to design the multimodal FL framework under semi-supervised setting where clients own private unlabeled data, and the server holds labeled data for global downstream tasks. To deal with the labeled data constraint of local clients in multimodal FL,~\citet{zhao2022multimodal} proposed a framework that works under the semi-supervised setting using multiple autoencoders for different modalities. This work applies the mechanism of the conventional federated averaging (FedAvg) framework~\citep{mcmahan2017communication} to construct the global multimodal encoders for supervised downstream tasks by aggregating model parameters from local autoencoders. However, these methods rely on aggregating model parameters from skewed private data on the server, which can lead to a decrease in the generalization ability of the global model due to statistical heterogeneity \citep{li2019fedmd, deng2020adaptive} and require a large \textit{communication overhead}. 

In this paper, we design a novel multimodal FL framework under a semi-supervised setting to tackle the limitations of existing multimodal FL works \citep{zhao2022multimodal,xiong2022unified} and resolve the labeled data constraint for FL clients while saving communication overhead. Specifically, we propose FedMEKT, a novel semi-supervised learning-based multimodal embedding knowledge transfer mechanism that exploits joint multimodal knowledge from unlabeled data of participating clients using proxy data to achieve the generalized encoder. %Acknowledging CreamFL~\cite{yu2023multimodal} as the first knowledge distillation-based multimodal federated learning (FL) framework, we distinguish our approach in several ways. Firstly, our approach operates in a semi-supervised setting, while CreamFL utilizes a supervised setting. Secondly, we propose a knowledge transfer scheme that leverages \textit{joint embedding knowledge} and performs distillation based on this joint representation. In contrast, CreamFL adopts a contrastive-based method for representation aggregation and incorporates separate representations for each modality during the local training process. These differences highlight the unique characteristics of our approach over CreamFL. \textcolor{red}{Aditionnally, many previous works on multimodal learning~\cite{} proved that the complementary information from different modalities can enhance the performance comparing to the unimodal feature.} 
We construct the FedMEKT algorithm for multimodal FL by exploiting the split multimodal autoencoder backbone from~\citep{ngiam2011multimodal}. This facilitates communication between the server and clients by leveraging a small multimodal proxy dataset. To handle the communication cost in multimodal FL, we develop a knowledge transfer mechanism that operates on both sides of the system, leveraging the joint knowledge of the learning models rather than relying on model parameters. Inspired by the effectiveness of joint embeddings generated from multimodal data, we raise the following research question: \textit{``how to effectively obtain joint embeddings for knowledge transfer between the server and multimodal heterogeneous clients?"}

To answer this question, we introduce a fusion layer to combine the knowledge generated from different modalities into joint embedding knowledge. Moreover, instead of updating the global model by aggregating local model parameters, our FedMEKT updates the generalized autoencoder via the joint embedding knowledge from participating clients. %Note that different from previous works \cite{liang2020think, liu2020federated} that generate the representation from private raw data, which may violate the privacy preservation in FL, our work adopts a multimodal proxy dataset to extract the personal knowledge that could not be reconstructed back to the original data, thus ensuring the privacy for local clients.
Unlike previous works~\citep{liang2020think, liu2020federated}, which generate representations directly from private raw data and may compromise privacy in FL, our approach leverages a multimodal proxy dataset. By extracting personal knowledge that cannot be reverse-engineered to reconstruct the original data, our method ensures privacy for local clients while preserving the efficacy of knowledge transfer. Through extensive experiments, we demonstrate that our proposed framework can achieve a significant performance gain and save communication costs compared to other methods on supervised tasks. Our main contributions are:
\begin{itemize}
     \item %We propose a knowledge transfer method for multimodal FL that can be specified as upstream and downstream mechanisms, and develop a novel multimodal FL framework FedMEKT that allows joint embedding knowledge exchange between the server and clients.
     We propose a novel upstream and downstream knowledge transfer method for multimodal FL, called FedMEKT. This method %incorporates upstream and downstream mechanisms, allowing for
     enables the exchange of joint multimodal embedding knowledge between the server and clients via a small proxy data %We propose %the downstream and upstream multimodal embedding knowledge transfer mechanisms between server and clients 
    considering the following problems: 1) \textit{local multimodal autoencoder learning} to update local multimodal encoders with the global multimodal knowledge received from the server, 2) \textit{generalized multimodal autoencoder construction} to aggregate the knowledge sent from heterogeneous client encoders to global encoders, and 3) \textit{generalized classifier learning} to train the classifier for global downstream tasks. 
    \item To facilitate the aggregation of joint multimodal embedding knowledge, we add a fusion layer and establish knowledge exchange between the server and clients, without the need for parameter exchange.
    \item We validate the proposed method by conducting extensive experiments over three multimodal human activity recognition datasets. As a result, out FedMEKT achieves superior performance in global downstream tasks while requiring far less communication overhead than other baselines.

\end{itemize}
\begin{figure*}[h]
	\centering
	\includegraphics[width=0.9\linewidth]{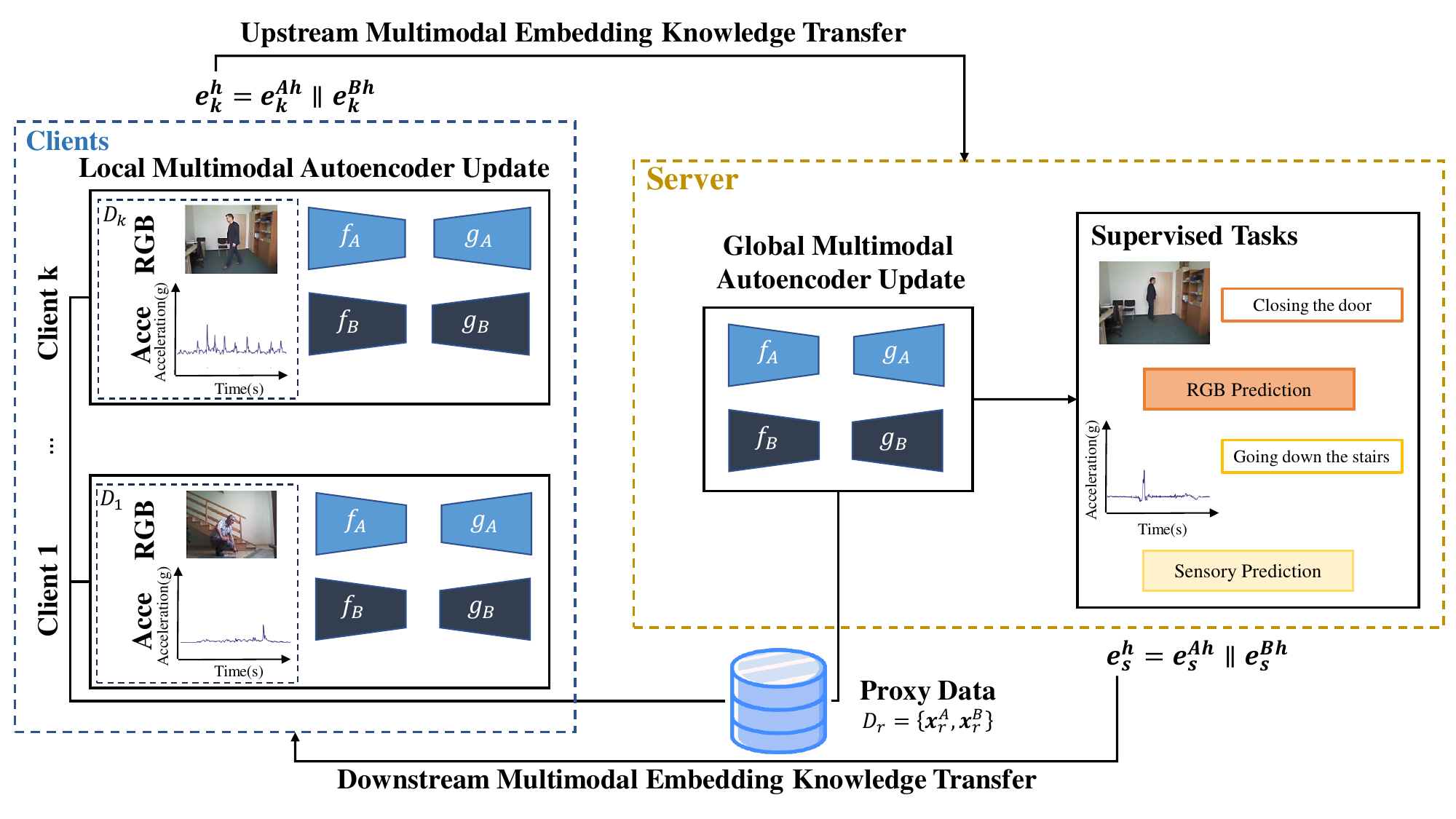}
	\caption{An overview of FedMEKT framework on UR Fall Detection dataset~\citep{kwolek2014human}. Multimodal clients update their models with private data. A proxy data is shared between the server and clients for the distillation step, and the server handles downstream tasks with labeled data.}
	\label{systemmodel}
\end{figure*}

\section{Related Work}

\paragraph{Semi-supervised Learning}  Semi-supervised learning (SSL) has been applied in various machine learning tasks to leverage the unlabeled data for solving the labeling cost issue, considering a scenario where the system holds both unlabeled and labeled data \citep{zhu2009introduction}. Pseudo-labeling \citep{lee2013pseudo} has become one of the popular trends in SSL, which generates the pseudo-label for unlabeled dataset and compute the loss function based on the sum of original and pseudo-label loss functions. The combination of pseudo-label and consistency regularization has been widely applied in SSL with many state-of-the-art methods, such as UDA \citep{xie2020unsupervised}, FixMatch~\citep{sohn2020fixmatch}, and MixMatch~\citep{berthelot2019mixmatch}. In our work, we consider a semi-supervised scenario in FL setting where private unlabeled data is provided by users, and labeled data is utilized in training classifiers to perform the downstream tasks. 

\paragraph{Knowledge Distillation based Federated Learning}  Knowledge distillation (KD) \citep{bucilua2006model,ba2013deep} has become a promising technique that helps FL to solve heterogeneity issues. KD generally provides knowledge communication between global and local models instead of exchanging model parameters. The authors in \citep{jeong2018communication,nguyen2022cdkt} applied KD in FL to minimize the communication overhead by using the distillation regularizer between student and teacher logits. The clients update their local models to conduct the averaged global predictions on the server. FedDF \citep{lin2020ensemble} first obtains a global model by aggregating the local model parameters and then updates it again by performing ensemble distillation from all client teacher models. Unlike FedDF, which still uses the model parameters exchange between the clients and the server, KT-pFL \citep{zhang2021parameterized} formulates the personalized knowledge transfer for personalized FL leveraging the proxy data to update the local soft predictions. However, most of these schemes apply KD in traditional FL with unimodal labeled data while we integrate the knowledge transfer scheme with the multimodal FL in this work.

\paragraph{Multimodal Learning}  Multimodal learning has attracted lots of attention in recent years. The multimodal deep learning systems enable leveraging data from multiple modalities, such as image \citep{bain2021frozen,ye2019cross}, video \citep{nagrani2021attention,bain2021frozen}, sensors \citep{islam2022mumu,caesar2020nuscenes}, etc., hence providing better performance than unimodal data. One of the first designs in multimodal deep learning was fusion \citep{potamianos2003recent,wollmer2010context} that fuses representations from different layers of multiple modalities in various ways, such as concatenation, multiplication, or weighted sum. However, these works still face misalignment in different fusion levels. In recent years, researchers have proposed different model architectures for multimodal applications such as co-attention \citep{lu2016hierarchical} for VQA tasks \citep{kumar2020mcqa}, and various types of transformers for language-video tasks~\citep{sun2019videobert}. In terms of the encoder-decoder framework, \citep{ngiam2011multimodal} proposed the autoencoders for audio and visual data. Another popular approach is DCCA \citep{andrew2013deep}, which is a combination of canonical correlation analysis (CCA) and autoencoders to fuse multimodal representations in the feature subspace. 

CreamFL~\citep{yu2023multimodal} is the first KD-based multimodal FL framework that adopts a contrastive-based method for representation aggregation and incoporates separate representations contrastive for each modality during the local training process. However, CreamFL works under a supervised setting, which may challenges the scarcity of training data and public data due to the high label annotation cost. Emerging multimodal FL works \citep{xiong2022unified,zhao2022multimodal} motivate us to initiate the novel design of a knowledge transfer scheme in multimodal FL. Our FedMEKT operates in a semi-supervised setting and performs the upstream and downstream knowledge transfer mechanisms. As a result, the global encoders could be built from local encoders by the joint multimodal embedding knowledge transfer and strengthen the performance of supervised tasks. In reverse, the local encoders utilize the embedding knowledge from the global encoder for enhancing their local learning ability. In the following section, we present our proposed Federated Multimodal Embedding Knowledge Transfer scheme and algorithm.
%------------------------------------------------------------------------
\begin{figure*}[h]
	\centering
	\includegraphics[width=0.85\linewidth]{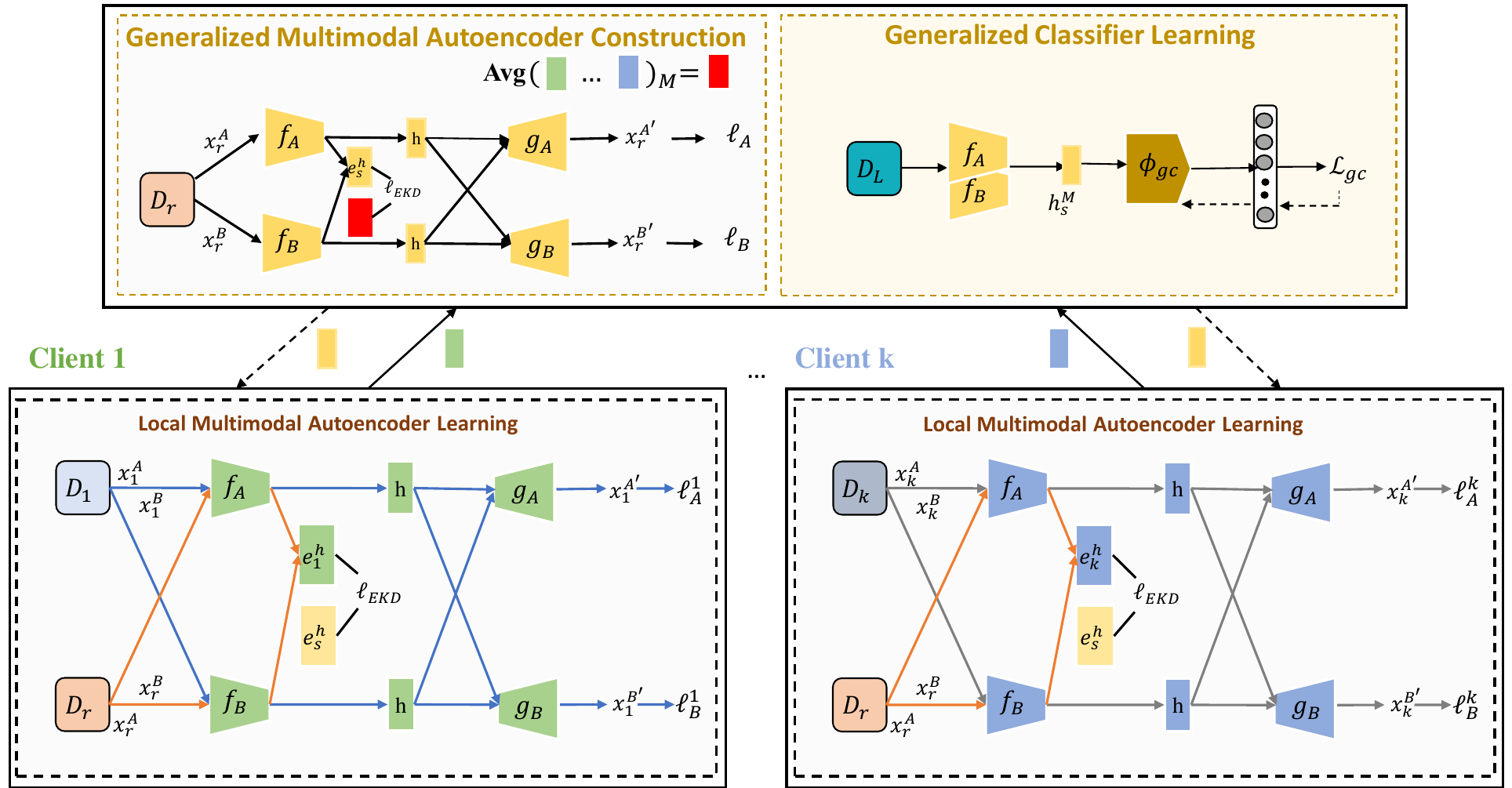}
	\caption{Illustration of FedMEKT considering three general problems: 1) local multimodal autoencoder learning updates the local models with global multimodal knowledge, 2) generalized multimodal autoencoder construction builds the global encoders with the joint knowledge distilled from clients, and 3) generalized classifier learning for downstream tasks. }
	\label{system_training_step}
\end{figure*}
\section{FedMEKT: Federated Multimodal Embedding Knowledge Transfer}
% \subsection{Preliminary}
A recent multimodal FedAvg scheme \citep{zhao2022multimodal} exploits multimodal data in FL by simply aggregating the parameters from local autoencoders trained on different data modalities. After the local training process with a private unlabeled dataset $D_k =\{\textbf{x}^M_k\}$ in client $k$, the global autoencoder is updated based on the aggregated model parameters from both unimodal and multimodal client models. The multimodal client models are given more weights than unimodal client models to align the representations from different modalities. 
\subsection{Problem Formulation}
Different from existing multimodal FL methods, which aggregate the parameters from clients in each communication round to train the global model, we develop the multimodal embedding knowledge transfer mechanism to transfer the local embedding knowledge from all multimodal clients to collectively build the generalized global encoder model $f_M$ which can generate more powerful representations for downstream tasks. Fig.~\ref{systemmodel} shows an overview of our FedMEKT framework on UR Fall Detection dataset~\citep{kwolek2014human}. Given two modalities $A$ and $B$, the corresponding autoencoders, $(f_{A},g_{A})$ and $(f_{B},g_{B})$, are used to minimize the distances between the original input data and the reconstructed data, which are referred as the reconstruction losses $\ell_A$ and $\ell_B$. In this work, we consider each client $k$ has either a multimodal dataset $D_k=\{\textbf{x}^A_k,\textbf{x}^B_k\}$ or a unimodal dataset $D_k=\{\textbf{x}^A_k\}$ or $D_k=\{\textbf{x}^B_k\}$, where multimodal clients own different types of sensor data (e.g., accelerometer data Acce and gyroscope data Gyro) or sensor and visual data at the same time (e.g., RGB images and accelerometer data Acce). Then, we construct a small unlabeled proxy dataset $D_r=\{\textbf{x}^A_r,\textbf{x}^B_r\}$ that all clients can access and provide their models' knowledge as the embedding of the samples in the proxy dataset. Note that this knowledge represents the learning capabilities of the models and is not necessarily matched to the true labels. In particular, we formulate the learning objective functions for our FedMEKT method as follow:
\begin{equation}
\begin{split}
     {\min_{w_{k}}{\mathcal{L}}^k_{c}= \min_{w_{k}}{\mathcal{L}}^k_{c}(f_{M},g_{M}|D_k,D_r),  \; \forall{M\in\{A,B\}},}
    \label{eq:objective-local}
\end{split}
\end{equation}
\begin{equation}
\begin{split}
     {\min_{w_{s}}{\mathcal{L}}_{s}= \min_{w_{s}}{\mathcal{L}}_{s}(f_{M},g_{M}|D_r),  \; \forall{M\in\{A,B\}},}
     \label{eq:objective-global}
\end{split}
\end{equation}
\begin{equation}
\begin{split}
     {\min_{w^M_{gc}}{\mathcal{L}}_{gc}= \min_{w^M_{gc}}{\mathcal{L}}_{gc}(\phi_{gc}|D_L),  \; \forall{M\in\{A,B\}},}
     \label{eq:objective-classifier}
\end{split}
\end{equation}
where $\mathcal{L}^k_c$, $\mathcal{L}_s$ and $\mathcal{L}_{gc}$ are the overall loss functions of local autoencoder, generalized autoencoder and generalized classifier models. The global encoder model shares its generalized joint embedding knowledge with all clients by upstream transfer and receives local joint embedding knowledge from clients through a downstream transfer mechanism using proxy data. This proxy dataset has a reasonable size and contains unlabeled data from all modalities that the system can collect easily with a low cost. Proxy data could be collected from the service provider which is out-of-distribution from all clients' local data.
%, and does not capture the full statistics of local data. 
It works as the bridge to deliver the knowledge from clients to servers and vice versa. In this design, the FL framework could guarantee better data privacy and protect model parameters compared to schemes that require parameter exchange between the server and clients. An illustration of our proposed FedMEKT scheme is shown in Fig.~\ref{system_training_step}. In the following subsection, we design the embedding knowledge transfer in multimodal FL for FedMEKT by formulating three general problems: \textit{local multimodal autoencoder learning}, \textit{generalized multimodal autoencoder construction},  and \textit{generalized classifier learning} in the following subsections.

\subsection{Local Multimodal Autoencoder Learning}
Our architecture aims to address the challenge of modality discrepancy among multiple clients on the local side. Since each client may have different modalities, designing a framework that can effectively learn the common features from multimodal data across all clients is crucial. To achieve this, we design the distillation-based multimodal knowledge transfer mechanism for local multimodal autoencoder learning on the client side, enabling the local models to improve the generalization capabilities by mimicking the global embedding knowledge $e^{h}_{s}$ received from the server during downstream multimodal embedding knowledge transfer. The embedding knowledge is generated from the encoder $f$ of modality $M$  (i.e., $e^M=f(x_r^M)$), and it could be extracted from different hidden layers of the encoder (i.e., $e^{Mh}$), where $h$ is the number of hidden layers of the encoder for the modality $M$. The embedding knowledge from modalities A and B are $e^{Ah}_k$ and $e^{Bh}_k$, respectively. To obtain the \textit{joint embedding knowledge}, we design the \emph{fusion layer} to concatenate the knowledge extracted from two modalities A and B, which is denoted as:
\begin{equation}
\begin{split}
     e^h_k=[e^{Ah}_k\mathbin\Vert e^{Bh}_k]
\end{split}
\end{equation}
The fusion layer has the same dimensions as the hidden layers of the encoder to store the joint embedding knowledge. In the local learning problem, the client models update depending on private unlabeled data $D_k$ with the local reconstruction loss and proxy data $D_r$ with embedding knowledge transfer loss ${\ell}_{EKD}$. In particular, we construct the local multimodal loss function for each device $k$ on both modalities A and B as follows:
\begin{algorithm}[t]
\footnotesize
\caption{FedMEKT Algorithm} 
\label{mainalg}
\begin{algorithmic}[1]
    \State \textbf{Input:} $T, R, N, P  $, $\eta_1$, $\eta_2$, $\eta_3$
    \For{ $t=0,\dots,T-1$}
\State -- \textit{\textbf{Client Execution}} --
    % \State \begin{varwidth}{\linewidth}
    
    \State \textbf{Local Multimodal Autoencoder Learning:}
    Clients receive  global joint embedding knowledge ${e}^{h}_{s}$ from the server 
    % \For{device $n=1,\dots,N  $}
    \For { $n=0,\dots,N-1$}
        % \State \begin{varwidth}[t]{0.85\linewidth}
        % We loop for each batch of private and proxy data (i.e., $S_k$ and $S_r$): \end{varwidth}
        \For {$S_k$ $\subseteq$~$D_k$, $S_r$ $\subseteq$~$D_r$ } 
            \begin{flalign*}
                &&&&&&{w}^{t}_{k} = &{w}^{t}_{k}-\eta_1\nabla \mathcal{L}_c^k  &&\text{\Comment{local model parameters ${w}_{k}$ update }}
            \end{flalign*}
         \State $\mathcal{L}_c^k$ is computed via Eq.~\ref{eq:totalloss-local} %\Comment{Local Autoencoder Update}
    \EndFor
    \EndFor
    \State \begin{varwidth}[t]{1.\linewidth}
    \end{varwidth}
    Device $k$ sends the joint embedding knowledge $e^{h}_{k}$ to the server using proxy data $D_r$ 
    % \end{varwidth}
    % \EndFor
    \State -- \textit{\textbf{Server Execution}} --
    %   \State \begin{varwidth}{\linewidth}
    \State \textbf{Generalized Multimodal Autoencoder Construction:} %Server distributes $e^h_s$ to all clients
    \For { $r=0,\dots,R-1$}
      \For {$S_r$ $\subseteq$~$D_r$ } 
      \begin{align*}
       &&&&&&{w}^{t}_{s}= &{w}^{t}_{s}-\eta_2\nabla\mathcal{L}_s  && \text{\Comment{global model parameters $w_s$ update}}
      \end{align*}
      \State $\mathcal{L}_s$ is computed via Eq.~\ref{eq:totalloss-server} %\Comment{Global Autoencoder Update}
     \EndFor
     \EndFor
    \State \textbf{Generalized Classifier Learning:} Global classifier receives $h_s^M$ from global encoders $\forall{M\in\{A,B\}}$
    \For { $p=0,\dots,P-1$}
      \For {$S_L$ $\subseteq$~$D_L$ } 
        \begin{align*}
              &&&&&&{w}^{t,M}_{gc}= &{w}^{t,M}_{gc}-\eta_3\nabla\mathcal{L}_{gc}   &&\text{\Comment{global classifier $w_{gc}$ update}}
        \end{align*}
      \State $\mathcal{L}_{gc}$ is computed via Eq.~\ref{eq:globalclassifier-mmEKT}%\Comment{Global Classifier Update} 
      \EndFor
    \EndFor 
    \EndFor
\end{algorithmic}
\end{algorithm}	
% \begin{equation}
%     {\ell}^{k}_{A}= {\ell}_{A}(x^A_k,x^{A'}_k|D_k) +{\ell}_{B}(x^B_k,x^{B'}_k|D_k)+\alpha \, {\ell}_{EKD}(e^{h}_{k},e^{h}_{s}|D_r)
%      \label{eq:ondeviceA-mmEKT}
% \end{equation}

\begin{equation}
\begin{split}
  {\mathcal{L}}^{k}_{A}= {\ell}_{A}(x^A_k,g_A(h_A)|D_k) &+{\ell}_{B}(x^B_k,g_B(h_A)|D_k) \\
  &+\gamma \, {\ell}_{EKD}(e^{h}_{k},e^{h}_{s}|D_r),
    \label{eq:ondeviceA-mmEKT}
\end{split}
\end{equation}
\begin{equation}
\begin{split}
  {\mathcal{L}}^{k}_{B}= {\ell}_{A}(x^A_k,g_A(h_B)|D_k) &+{\ell}_{B}(x^B_k,g_B(h_B)|D_k) \\
  &+\gamma \, {\ell}_{EKD}(e^{h}_{k},e^{h}_{s}|D_r),
     \label{eq:ondeviceB-mmEKT}
\end{split}
\end{equation}
where $h_A=f_A(x^A_k)$, $h_B=f_B(x^B_k)$, and $\gamma$ is the parameter to manipulate the trade-off between the local reconstruction loss and the embedding knowledge transfer regularizer. $\ell_A$ and $\ell_B$ are the reconstruction loss functions of two corresponding modalities A and B on local data, respectively. Hence, the $\ell_{EKD}$ denotes the joint embedding knowledge transfer regularizer (e.g., Kullback-Leibler (KL) Divergence loss) for both modalities A and B. By mimicking the generalized encoder via minimizing the $\ell_{EKD}$, the local model could enhance the generalization of local multimodal encoder models and avoid biases when training on the skewed private dataset. If client $k$ holds the unimodal data of modality A or B, it is only necessary to compute the autoencoder loss function for that particular modality. To perform the local autoencoder learning, we apply the synchronous update for the entire autoencoder from both modalities. The overall training loss $\mathcal{L}^k_c$ defined in Eq.~\ref{eq:objective-local} for the autoencoders from two modalities for each client $k$ is given in Eq.~\ref{eq:totalloss-local}.
\begin{equation}
     \mathcal{L}^k_c=\mathcal{L}^k_A+\mathcal{L}^k_B.
     \label{eq:totalloss-local}
\end{equation}

\subsection{Generalized Multimodal Autoencoder Construction}
% To solve the generalized multimodal autoencoder construction problem, we utilize the proxy data $D_r$ to collect the embedding knowledge from all multimodal devices to build the global autoencoder model. On the server, we conduct the modality-averaging mechanism of the local embedding knowledge based on the modality label. Accordingly, we gather the embedding knowledge from same modality of all clients and then perform the averaging operation to obtain the collective knowledge of each modality for the global model. Subsequently, we design the generalized multimodal autoencoder construction problem for the split autoencoder with two modalities using embedding knowledge distillation (EKD) \cite{}method as follow:
%  We define that the embedding knowledge is generated from the encoder $f$ of modality $M$  (i.e., $e^M=f(x_r^M)$) and it could be extracted from different hidden layers of the encoder (i.e., $e^{M_h}$), where $h$ is the number of hidden layers of the encoder for the modality $M$.
On the server side, we target to perform joint knowledge aggregation in order to capture the biased features of the local models and learn a generalized global encoder for downstream tasks. To solve the generalized multimodal autoencoder construction problem, we utilize the proxy data $D_r$ to generate the global multimodal embedding knowledge from the current global model and collect the multimodal embedding knowledge from local clients to perform the upstream multimodal embedding knowledge transfer. Firstly, we concatenate the global embedding knowledge from different modalities to achieve the joint global embedding knowledge $e^h_s$. %For the local embedding knowledge, we first conduct the modality-averaging mechanism of the local embedding knowledge based on the modality label. 
Accordingly, we gather the embedding knowledge from the same modality of all clients and then perform the averaging operation to obtain the collaborative embedding knowledge of each modality, which are $\sum_{k\in{K}}{\frac{1}{K}}e^{Ah}_{k}$ and $\sum_{k\in{K}}{\frac{1}{K}}e^{Bh}_{k}$. Hence, similar to the client side, we concatenate the \textit{collaborative embedding knowledge} from two modalities, A and B, by using the fusion layer and achieve the joint local embedding knowledge, which is denoted as:
\begin{equation}
\begin{split}
    \sum_{k\in{K}}{\frac{1}{K}}e^{h}_{k}=[\sum_{k\in{K}}{\frac{1}{K}}e^{Ah}_{k}\mathbin\Vert \sum_{k\in{K}}{\frac{1}{K}}e^{Bh}_{k}]
\end{split}
\end{equation}
The global autoencoder model is updated with reconstruction loss and global embedding knowledge transfer loss using proxy data $D_r$. Subsequently, we design the global multimodal learning loss for two modalities as follows:

\begin{equation}
\begin{split}
       {\mathcal{L}}^s_{A}= {\ell}_{A}(x^A_r,g_A(h_A)|D_r) & +{\ell}_{B}(x^B_r,g_B(h_A)|D_r) \\
       &+\beta \, {\ell}_{EKD}\bigg(e^{h}_{s},\sum_{k\in{K}}{\frac{1}{K}}e^{h}_{k}|D_r\bigg),
       \label{eq:globalA-mmEKT}
\end{split}
\end{equation}

\begin{equation}
\begin{split}
    {\mathcal{L}}^s_{B}= {\ell}_{A}(x^A_r,g_A(h_B)|D_r)& +{\ell}_{B}(x^B_r,g_B(h_B)|D_r) \\
    &+\beta \, {\ell}_{EKD}\bigg(e^{h}_{s},\sum_{k\in{K}}{\frac{1}{K}}e^{h}_{k}|D_r\bigg),
        \label{eq:globalB-mmEKT}
\end{split}
\end{equation}
where $h_A=f_A(x^A_r)$, $h_B=f_B(x^B_r)$, and $\beta$ is the parameter to control the trade-off between reconstruction loss of proxy data and embedding knowledge transfer regularizer. Here, $\ell_A$ and $\ell_B$ are the reconstruction loss functions of two corresponding modalities A and B on proxy data, respectively. The embedding knowledge transfer regularizer (i.e., $\ell_{EKD}$) attempts to close the gap between the multimodal embedding knowledge of the server and joint embedding knowledge from multimodal clients.  Same as the client side, we leverage the synchronous update for the generalized multimodal autoencoder model. The total loss Eq.~\ref{eq:objective-global}  for global multimodal autoencoder construction can be defined as follows: 
\begin{equation}
     \mathcal{L}_s=\mathcal{L}^s_A+\mathcal{L}^s_B.
     \label{eq:totalloss-server}
\end{equation}

\subsection{Generalized Classifier Learning}
On the server, we attach the global classifier $\phi_{gc}$ to global encoder part of the global autoencoder model to perform the supervised learning task by using multimodal labeled dataset $D_L$. The global training process helps to update solely the global classifier for classification downstream tasks. Hence, we use cross-entropy loss to learn the global multimodal classifier and solve the problem in Eq.~\ref{eq:objective-classifier}:

\begin{equation}
    {\mathcal{L}}_{gc}= {\mathcal{L}}_{CE}(z^M_s|D_L), \; \forall{M\in\{A,B\}}, 
    \label{eq:globalclassifier-mmEKT}
\end{equation}
where ${\mathcal{L}}_{CE}$ is the cross-entropy loss and $h^{M}_{s}=f_{M}(x^{M}_L)$, $z^M_s=\phi_{gc}(h^{M}_{s})$ are the representations from the global encoder and the outcome of the global classifier, respectively. 
\subsection{FedMEKT Algorithm}
Turning this multimodal FL scheme into reality, we develop the FedMEKT algorithm (Alg.~\ref{mainalg}) to perform the embedding knowledge transfer between the server and multimodal clients. At the beginning of each communication round, the server randomly selects a subset of clients from the total of $K$ multimodal clients to participate in the local training, and the global autoencoder model broadcasts the generalized embedding knowledge to all selected clients. Each client performs $N$ local training steps with its private multimodal data and proxy dataset in the local knowledge transfer problem~(Eq.~\ref{eq:totalloss-local}) and outputs the local embedding knowledge using proxy data $D_r$, then sends it to the server. The generalized autoencoder model is constructed on the server side by solving the generalized multimodal learning problem~(Eq.~\ref{eq:totalloss-server}). Then, the server uses the updated global encoder to extract the multimodal representations of input data in the labeled dataset $D_L$ to train the classifier for the supervised learning task (Eq.~\ref{eq:globalclassifier-mmEKT}). 

\section{Experimental Results}
% \subsection{Experiment Details}
% \subsubsection{Dataset}

\begin{table}[h]
\caption{Architecture details of the multimodal autoencoder used in our experiments for different datasets.}
\begin{center}
\begin{tabular}{ccccc}
\hline
\textbf{Dataset} & \textbf{Modality}                                          & \textbf{X size}                                     & \textbf{$h_1$ size} & \textbf{$h_2$ size} \\ \hline
mHealth        & \begin{tabular}[c]{@{}c@{}}Acce\\ Gyro\\ Mage\end{tabular} & \begin{tabular}[c]{@{}c@{}}9\\ 6\\ 6\end{tabular}   & 4                & 24               \\ \hline
UR Fall         & \begin{tabular}[c]{@{}c@{}}Acce\\ RGB\\ Depth\end{tabular} & \begin{tabular}[c]{@{}c@{}}3\\ 512\\ 8\end{tabular} & 2,4              & 32               \\ \hline
Opp             & \begin{tabular}[c]{@{}c@{}}Acce\\ Gyro\end{tabular}        & \begin{tabular}[c]{@{}c@{}}24\\ 15\end{tabular}     & 10               & 24               \\ \hline
\end{tabular}

\label{detai_model}
\end{center}
\end{table}

\begin{table*}[h]
\caption{Overall information of the multimodal autoencoder architecture used in our experiments.}
\begin{center}
\begin{tabular}{@{}ll|lll@{}}
\toprule
\multicolumn{2}{c|}{\multirow{2}{*}{\textbf{Components}}} & \multicolumn{3}{c}{\multirow{2}{*}{\textbf{Layer Detail}}} \\
\multicolumn{2}{c|}{}                                     & \multicolumn{3}{c}{}                                             \\ \midrule
\multicolumn{2}{l|}{\multirow{2}{*}{Encoder A}}           & \multicolumn{3}{l}{(lstm): LSTM(input\_layer A, hidden\_layer 1A, batch\_first=True)}       \\
\multicolumn{2}{l|}{}                                     & \multicolumn{3}{l}{(lstm2): LSTM(hidden\_layer 1A, hidden\_layer 2A, batch\_first=True)}      \\ \midrule
\multicolumn{2}{l|}{\multirow{2}{*}{Decoder A}}           & \multicolumn{3}{l}{(lstm): LSTM(hidden\_layer 2A, hidden\_layer 1A, batch\_first=True)}       \\
\multicolumn{2}{l|}{}                                     & \multicolumn{3}{l}{(lstm2): LSTM(hidden\_layer 1A, output\_layer A, batch\_first=True)}      \\ \midrule
\multicolumn{2}{l|}{\multirow{2}{*}{Encoder B}}           & \multicolumn{3}{l}{(lstm): LSTM(input\_layer B, hidden\_layer 1B, batch\_first=True)}     \\
\multicolumn{2}{l|}{}                                     & \multicolumn{3}{l}{(lstm2): LSTM(hidden\_layer 1B, hidden\_layer 2B, batch\_first=True)}      \\ \midrule
\multicolumn{2}{l|}{\multirow{2}{*}{Decoder B}}           & \multicolumn{3}{l}{(lstm): LSTM(hidden\_layer 2B, hidden\_layer 1B, batch\_first=True)}       \\
\multicolumn{2}{l|}{}                                     & \multicolumn{3}{l}{(lstm2): LSTM(hidden\_layer 1B, output\_layer B, batch\_first=True)} \\ \bottomrule  
\end{tabular} 
\label{table:model_architecture}
\end{center}
\end{table*}

\begin{table*}[t]
\caption{The comparison of average accuracy on different datasets with multimodal clients.  }
\begin{center}
\resizebox{17cm}{!}{%
\begin{tabular}{|c|cc|c|c|c|c|c|c|}
\hline
Datasets                 & \multicolumn{2}{c|}{Modality}                            & MM-FedAvg & MM-FedProx & MM-MOON & CreamFL        & FedMEKT-S & FedMEKT-C      \\ \hline
\multirow{6}{*}{mHealth} & \multicolumn{1}{c|}{\multirow{2}{*}{Acce-Gyro}}  & Acce  & 63.83     & 67.86      & 64.47   & \textbf{68.58}          & 64.13     & 68.28 \\ 
                         & \multicolumn{1}{c|}{}                            & Gyro  & 63.62     & 64.36      & 64.12   & 64.34          & 62.10     & \textbf{64.60} \\ \cline{2-3} 
                         & \multicolumn{1}{c|}{\multirow{2}{*}{Acce-Mage}}  & Acce  & 69.99     & 70.00      & 69.71   & 69.51          & 70.15     & \textbf{71.16} \\ 
                         & \multicolumn{1}{c|}{}                            & Mage  & 68.49     & 69.21      & 68.79   & 69.18          & 70.12     & \textbf{71.13} \\ \cline{2-3} 
                         & \multicolumn{1}{c|}{\multirow{2}{*}{Gyro-Mage}}  & Gyro  & 65.43     & 65.37      & 65.90   & 66.05          & 64.47     & \textbf{67.10} \\ 
                         & \multicolumn{1}{c|}{}                            & Mage  & 68.28     & 68.75      & 68.13   & 68.00          & 67.82     & \textbf{69.02} \\ \hline
\multirow{6}{*}{UR-Fall} & \multicolumn{1}{c|}{\multirow{2}{*}{Acce-RGB}}   & Acce  & 61.70     & 65.66      & 65.89   & 66.82          & 69.32     & \textbf{70.66} \\  
                         & \multicolumn{1}{c|}{}                            & RGB   & 57.88     & 59.26      & 62.00   & 62.34          & 60.21     & \textbf{66.70} \\ \cline{2-3} 
                         & \multicolumn{1}{c|}{\multirow{2}{*}{Acce-Depth}} & Acce  & 67.24     & 68.79      & 68.08   & 71.61          & 69.25     & \textbf{72.68} \\  
                         & \multicolumn{1}{c|}{}                            & Depth & 60.76     & 68.08      & 68.16   & \textbf{75.33} & 65.85     & 75.22          \\ \cline{2-3} 
                         & \multicolumn{1}{c|}{\multirow{2}{*}{RGB-Depth}}  & RGB   & 69.88     & 75.60      & 73.38   & \textbf{78.57} & 73.81     & 77.87          \\ 
                         & \multicolumn{1}{c|}{}                            & Depth & 67.61     & 70.04      & 66.48   & 69.18          & 70.33     & \textbf{70.57} \\ \hline
\multirow{2}{*}{Opp}     & \multicolumn{1}{c|}{\multirow{2}{*}{Acce-Gyro}}  & Acce  & 71.75     & 72.25      & 72.96   & 73.33          & 72.03     & \textbf{73.50} \\ 
                         & \multicolumn{1}{c|}{}                            & Gyro  & 72.08     & 71.43      & 72.09   & 72.10          & 72.09     & \textbf{72.15} \\ \hline
\end{tabular}
}%
\end{center}
\label{multimodal_table}
\end{table*}

\begin{table*}[t]
\caption{The comparison of average accuracy on different datasets with mixed clients. }
% \footnotesize
\begin{center}
\resizebox{17cm}{!}{%
\begin{tabular}{|c|cc|c|c|c|c|c|c|}
\hline
Datasets                 & \multicolumn{2}{c|}{Modality}                            & MM-FedAvg & MM-FedProx & MM-MOON & CreamFL       & FedMEKT-S      & FedMEKT-C      \\ \hline
\multirow{6}{*}{mHealth} & \multicolumn{1}{c|}{\multirow{2}{*}{Acce-Gyro}}  & Acce  & 64.19     & 65.27      & 67.97   & 66.89          & 64.32          & \textbf{68.37} \\ 
                         & \multicolumn{1}{c|}{}                            & Gyro  & 63.50     & 64.37      & 64.76   & 64.72          & 64.06          & \textbf{65.19} \\ \cline{2-3} 
                         & \multicolumn{1}{c|}{\multirow{2}{*}{Acce-Mage}}  & Acce  & 68.41     & 66.29      & 66.33   & 67.78          & 67.96          & \textbf{68.71} \\  
                         & \multicolumn{1}{c|}{}                            & Mage  & 68.88     & 67.76      & 68.67   & 69.52          & 69.11          & \textbf{70.68} \\ \cline{2-3} 
                         & \multicolumn{1}{c|}{\multirow{2}{*}{Gyro-Mage}}  & Gyro  & 65.57     & 66.30      & 64.65   & 66.68          & 65.09          & \textbf{67.38} \\ 
                         & \multicolumn{1}{c|}{}                            & Mage  & 68.14     & 68.59      & 68.50   & 69.52          & \textbf{69.68} & 69.03          \\ \hline
\multirow{6}{*}{UR-Fall} & \multicolumn{1}{c|}{\multirow{2}{*}{Acce-RGB}}   & Acce  & 73.60     & 69.63      & 64.61   & 69.72          & 67.69          & \textbf{73.92} \\  
                         & \multicolumn{1}{c|}{}                            & RGB   & 61.61     & 62.45      & 64.66   & \textbf{68.28} & 60.42          & 68.18          \\ \cline{2-3} 
                         & \multicolumn{1}{c|}{\multirow{2}{*}{Acce-Depth}} & Acce  & 62.18     & 73.45      & 70.74   & \textbf{75.24} & 68.88          & 73.70          \\ 
                         & \multicolumn{1}{c|}{}                            & Depth & 71.38     & 68.82      & 71.91   & \textbf{72.73} & 66.87          & 71.97          \\ \cline{2-3} 
                         & \multicolumn{1}{c|}{\multirow{2}{*}{RGB-Depth}}  & RGB   & 66.99     & 74.27      & 77.72   & 76.51          & 75.75          & \textbf{80.13} \\ 
                         & \multicolumn{1}{c|}{}                            & Depth & 66.21     & 67.48      & 70.97   & 68.38          & 72.48          & \textbf{74.10} \\ \hline
\multirow{2}{*}{Opp}     & \multicolumn{1}{c|}{\multirow{2}{*}{Acce-Gyro}}  & Acce  & 70.59     & 71.81      & 70.84   & 72.33          & 72.14          & \textbf{73.61} \\ 
                         & \multicolumn{1}{c|}{}                            & Gyro  & 71.12     & 72.10      & 72.09   & 72.11          & 72.09          & \textbf{72.13} \\ \hline
\end{tabular}
}%
\end{center}
\label{mixed_table}
\end{table*}
\subsection{Experimental Setup}
\subsubsection{Datasets} In this section, we evaluate the efficiency of the FedMEKT algorithm using three multimodal human activity recognition (HAR) datasets: mHealth~\citep{banos2014mhealthdroid}, UR Fall Detection~\citep{kwolek2014human}, and Opportunity (Opp)~\citep{chavarriaga2013opportunity}. To conduct the experiments, we randomly select $10$ clients from a total of $30$ clients in each round. We follow the similar setting of MM-FedAvg~\citep{zhao2022multimodal} to generate the training and test data for the federated systems for all three datasets. For the proxy dataset of each dataset, we generate a subset of total data with a size approximately equal to the testing data. For supervised training, we randomly sample labeled data from the training dataset. In each dataset, we use two modalities as the input modalities for the training process. %More details about the datasets are provided in the Appendix.
\paragraph{mHealth~\citep{banos2014mhealthdroid}}This dataset consists of the on-body sensor records from $10$ participants doing $13$ daily living and exercise activities. In this dataset, we consider three modalities, accelerometer (Acce), gyroscope (Gyro), and magnetometer (Mage), in our experiments. We randomly use data from one participant as testing data, another participant's data as proxy data, and the rest as training data.
\paragraph{UR Fall Detection~\citep{kwolek2014human}}This dataset contains $70$ video clips of human activities with $3$ classes. We consider $3$ modalities RGB camera (RGB), depth camera (Depth), and sensory data of each video frame from accelerometers (Acce) in our experiments. We follow the setup from MM-FedAvg~\citep{zhao2022multimodal} to generate RGB data. We randomly sample $1/10$ of data as testing data, $1/10$ of data as proxy data, and the rest as training data. 
\paragraph{Opportunity~\citep{chavarriaga2013opportunity}}This is the human activity recognition dataset that consists of the on-body sensor records from $4$ participants doing kitchen activities. For each participant, they recorded five different runs. In this dataset, we utilize 18 kitchen activities as $18$ classes and consider two modalities, accelerometers (Acce) and gyroscope (Gyro) for our experiments. We use runs $4$ and $5$ from participants $2$ and $3$ as the testing data, runs $4$ and $5$ from participants $1$ and $4$ as proxy data, and the rest for training data, respectively. 
\subsubsection{Model Architecture and Baselines}
To illustrate the effectiveness of the joint embedding knowledge in knowledge transfer based multimodal FL, we provide the split knowledge variant of FedMEKT. Overall, FedMEKT comprises two versions, FedMEKT-C and FedMEKT-S, with FedMEKT-C being the primary algorithm that utilizes joint multimodal knowledge for knowledge transfer. In contrast, FedMEKT-S employs split knowledge from each modality, updating autoencoders of modalities A and B separately with the respective modality's split knowledge instead of concatenating the knowledge of different modalities from all clients as FedMEKT-C does. In Tables~\ref{detai_model} and~\ref{table:model_architecture}, we provide a summary of the autoencoder architecture that we employed in this paper. We follow the approach taken by \citet{zhao2022multimodal} and provide the model size for each dataset. In contrast to the prior work, we use two hidden layers as primary layers and extract knowledge from those layers for both upstream and downstream transfer mechanisms. This allows us to effectively leverage the learned representations for a wide range of tasks. In this study, we compared our proposed method, FedMEKT, with several baselines, including MM-FedAvg~\citep{zhao2022multimodal}, CreamFL~\citep{yu2023multimodal} and multimodal versions of state-of-the-art approaches, i.e., FedProx \citep{li2020federated} and MOON \citep{li2021model}.

\subsubsection{Implementation Details} The FedMEKT algorithm was developed using the Pytorch library \citep{paszke2019pytorch} and experiments were simulated on a server with one NVIDIA GeForce GTX-1080 Ti GPU using CUDA version 11.2 and an Intel Core i7-7700K 4.20GHz CPU with sufficient memory for model training. The knowledge transfer scheme utilized LSTM \citep{hochreiter1997long} autoencoders with 2 LSTM layers, and knowledge was extracted from 2 hidden layers. For downstream tasks, the global classifier was implemented as a two-layer perceptron with ReLU~\citep{nair2010rectified} activation function. 

\subsubsection{Evaluation Metrics} We evaluate the performance of the global encoder by extracting the representations for downstream tasks. Specifically, we freeze the parameters of the global encoder and train a linear classifier on the extracted representations. We report the average $F_1$ score in last 10 communication rounds as our evaluation metric.

\subsection{Experimental Results}

\subsubsection{Performance Comparison}

\begin{figure}[h]
\centering
	\includegraphics[width=\linewidth]{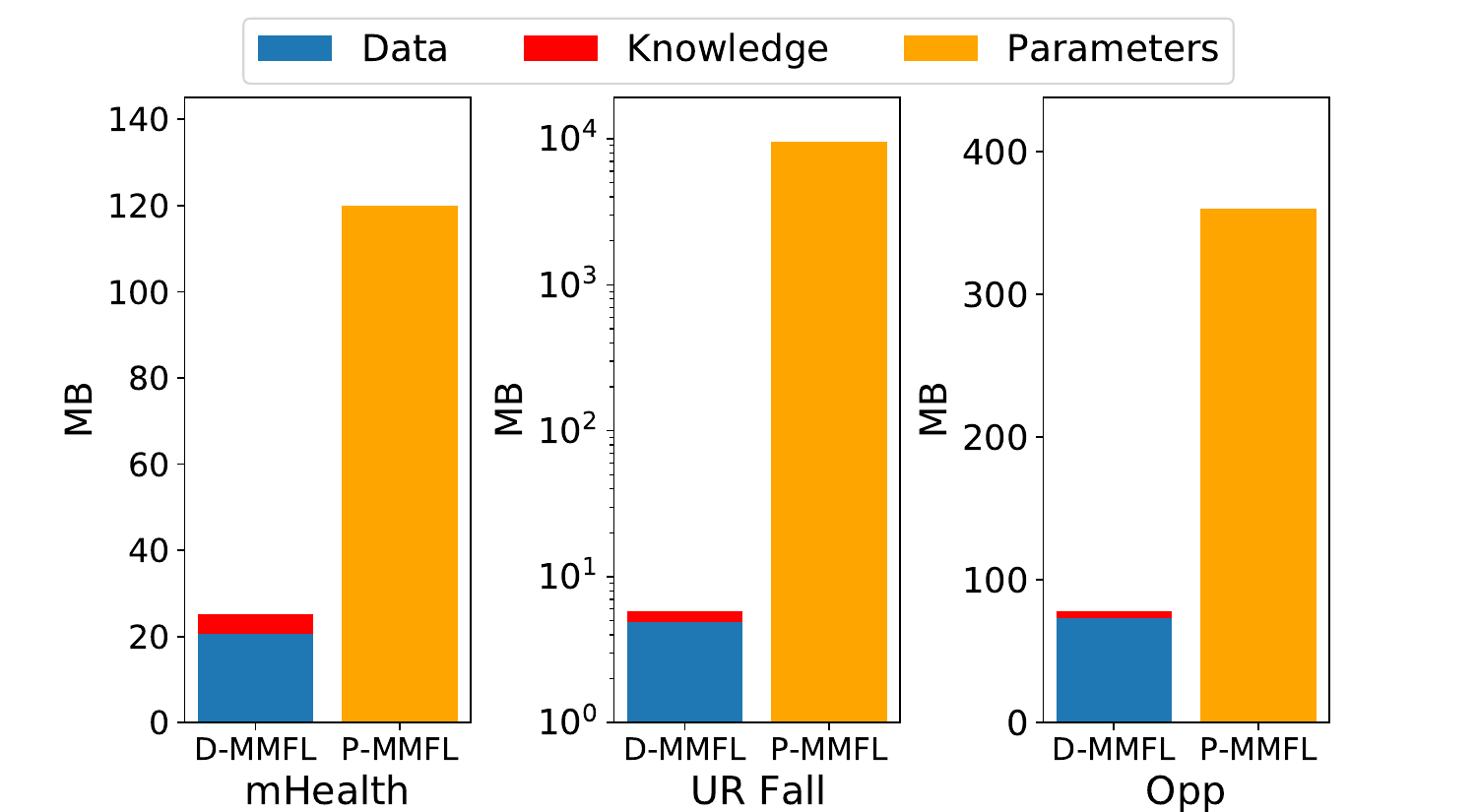}
    \caption{Communication Efficiency on three datasets. D-MMFL: distillation-based multimodal federated learning; P-MMFL: parameter-based multimodal federated learning. In this experiment, we evaluate with 10 participating clients in 100 communication rounds for all datasets.}
    \label{cost}
\end{figure}

\paragraph{Multimodal Clients} Table~\ref{multimodal_table} presents the performance comparison of our FedMEKT algorithm and other baselines on three multimodal human activity recognition datasets with $30$ multimodal clients. As the results show, FedMEKT-C achieves the highest test accuracy in most cases on the mHealth dataset with approximately $\textbf{0.5-2\%}$ improvement over other baselines. Similarly, on the UR Fall Detection dataset, FedMEKT-C also outperforms other baselines with $\textbf{1-2\%}$ improvement in most modalities combinations. While our method may not surpass CreamFL in certain combination cases, we still achieve competitive results. On the Opp dataset, FedMEKT-C obtains the highest accuracy of $\textbf{73.50\%}$ followed by CreamFL with the test accuracy of $\textbf{73.33\%}$ on the Acce downstream task. We observe that FedMEKT-S and MM-FedProx also achieve relatively high test accuracies in some cases but still slightly worse than FedMEKT-C. Using the embedding knowledge transfer method instead of exchanging model parameters, FedMEKT prevents reverse engineering and saves communication costs when the model size is large. The details will be introduced in the later experiment as shown in Fig.~\ref{cost}. Overall, the results demonstrate that our FedMEKT method, especially the FedMEKT-C variant, outperforms all baselines across all modalities combinations and datasets, indicating its effectiveness in addressing the challenges of multimodal federated learning. %We claim that FedMEKT-C has the best performance in most cases as the global model can adaptively learn better from the common representations derived from the joint knowledge of all clients than from learning from the split knowledge. 
In the majority of cases, FedMEKT-C demonstrates superior performance as the global model effectively learns from the joint embedding knowledge of all clients, encompassing common representations, compared to learning from the split knowledge.

\begin{table}[t]
\caption{The comparison of average accuracy of FedMEKT under different proxy data size on UR Fall Detection Dataset~\citep{kwolek2014human} with multimodal clients. }
% \footnotesize
\begin{center}
\begin{tabular}{cc|ccc|}
\hline
\multicolumn{2}{|c|}{\multirow{2}{*}{ Modality    }}             & \multicolumn{3}{c|}{Size of Proxy Data $D_r$}                                        \\ \cline{3-5}
\multicolumn{2}{|c|}{}                                    & \multicolumn{1}{c}{10\%}   & \multicolumn{1}{c}{50\%}            & 100\%  \\ \hline
\multicolumn{1}{|c|}{\multirow{2}{*}{Acce-RGB}}   & Acce  & \multicolumn{1}{c}{58.05} & \multicolumn{1}{c}{61.16}          & \textbf{70.66} \\ 
\multicolumn{1}{|c|}{}                            & RGB   & \multicolumn{1}{c} {58.27} & \multicolumn{1}{c}{60.43}          & \textbf{66.70} \\  \cline{1-2}
\multicolumn{1}{|c|}{\multirow{2}{*}{Acce-Depth}} & Acce  & \multicolumn{1}{c}{59.27} & \multicolumn{1}{c}{64.39}          & \textbf{72.68} \\  
\multicolumn{1}{|c|}{}                            & Depth & \multicolumn{1}{c}{58.22} & \multicolumn{1}{c}{62.93}          & \textbf{75.22} \\  \cline{1-2}
\multicolumn{1}{|c|}{\multirow{2}{*}{RGB-Depth}}  & RGB   & \multicolumn{1}{c}{67.43} & \multicolumn{1}{c}{72.99}          & \textbf{77.87} \\
\multicolumn{1}{|c|}{}                            & Depth & \multicolumn{1}{c}{56.88} & \multicolumn{1}{c}{\textbf{72.55}} & 70.57    \\ \hline
\end{tabular}
\label{multimodal_proxy_data}
\end{center}
\end{table}

\begin{table}[t]
\caption{The comparison of average accuracy of FedMEKT under different proxy data size on UR Fall Detection Dataset~\citep{kwolek2014human} with mixed clients. }
% \footnotesize
\begin{center}
\begin{tabular}{cc|ccc|}
\hline
\multicolumn{2}{|c|}{\multirow{2}{*}{ Modality    }}             & \multicolumn{3}{c|}{Size of Proxy Data $D_r$}                                        \\ \cline{3-5}
\multicolumn{2}{|c|}{}                                    & \multicolumn{1}{c}{10\%}   & \multicolumn{1}{c}{50\%}            & 100\%  \\ \hline
\multicolumn{1}{|c|}{\multirow{2}{*}{Acce-RGB}}   & Acce  & \multicolumn{1}{c}{63.40} & \multicolumn{1}{c}{65.02}          & \textbf{73.92} \\ 
\multicolumn{1}{|c|}{}                            & RGB   & \multicolumn{1}{c} {58.22} & \multicolumn{1}{c}{60.53}          & \textbf{68.18} \\  \cline{1-2}
\multicolumn{1}{|c|}{\multirow{2}{*}{Acce-Depth}} & Acce  & \multicolumn{1}{c}{58.43} & \multicolumn{1}{c}{61.95}          & \textbf{73.70} \\  
\multicolumn{1}{|c|}{}                            & Depth & \multicolumn{1}{c}{58.22} & \multicolumn{1}{c}{62.88}          & \textbf{71.97} \\  \cline{1-2}
\multicolumn{1}{|c|}{\multirow{2}{*}{RGB-Depth}}  & RGB   & \multicolumn{1}{c}{62.89} & \multicolumn{1}{c}{67.92}          & \textbf{80.13} \\
\multicolumn{1}{|c|}{}                            & Depth & \multicolumn{1}{c}{58.16} & \multicolumn{1}{c}{60.15} & \textbf{74.10}    \\ \hline
\end{tabular}
\label{mixed_proxy_data}
\end{center}
\end{table}

\paragraph{Mixed Clients} Apart from the multimodal clients setting, we extend the experiment to include a mixed clients setting with both unimodal and multimodal clients. To conduct this experiment, we include $10$ multimodal clients, $10$ unimodal clients for modality A, $10$ unimodal clients for modality B, and randomly select $10$ clients from a total of $30$ mixed clients in each communication round. Following \citep{zhao2022multimodal}, parameter-based methods such as MM-FedAvg, MM-FedProx, and MM-MOON assigned higher weights (set to 100) to multimodal clients in model aggregation. For FedMEKT variants like FedMEKT-C and FedMEKT-S with unimodal clients, we generate the knowledge from the encoder of the corresponding modality and obtain the collaborative knowledge of each modality by averaging the knowledge from both multimodal and unimodal clients. Table~\ref{mixed_table} shows the performance comparison of our methods with other baselines on downstream tasks in mixed clients setting. For most modalities combinations of three datasets, FedMEKT-C outperforms other baselines on downstream tasks performance. Exceptions include the Mage task in the Gyro-Mage combination of the mHealth dataset and certain combination cases in the UR Fall Detection dataset, where FedMEKT-S and CreamFL show slightly better performance. Nonetheless, we still achieve competitive results in these cases. From our observation, FedMEKT outperforms other baselines with $\textbf{1-3\%}$ in most combination cases, and some cases can improve the performance of downstream tasks when working on mixed clients setting. Overall, from the results in Table~\ref{mixed_table}, we claim that our proposed method can work well in various practical scenarios and outperforms both multimodal versions of other classical FL methods and state-of-the-art methods such as MM-FedAvg and CreamFL. Furthermore, the results clearly demonstrate that transferring complementary information from various modalities enhances performance significantly, surpassing traditional split knowledge transfer approaches like FedMEKT-S and CreamFL.

\begin{table}[t]
\caption{The comparison of average accuracy of FedMEKT under different EKT Steps on UR Fall Detection Dataset~\citep{kwolek2014human} with multimodal clients. }
\begin{center}
\begin{tabular}{|cc|ccc|}
\hline
\multicolumn{2}{|c|}{\multirow{2}{*}{Modality}}                   & \multicolumn{3}{c|}{\# of EKT Steps R}                                   \\ \cline{3-5} 
\multicolumn{2}{|c|}{}                                    & \multicolumn{1}{c}{1}     & \multicolumn{1}{c}{2}              & 3     \\ \hline
\multicolumn{1}{|c|}{\multirow{2}{*}{Acce-RGB}}   & Acce  & \multicolumn{1}{c}{53.73} & \multicolumn{1}{c}{\textbf{70.66}} & 66.28 \\  
\multicolumn{1}{|c|}{}                            & RGB   & \multicolumn{1}{c}{58.25} & \multicolumn{1}{c}{\textbf{66.70}} & 58.72 \\ \cline{1-2}
\multicolumn{1}{|c|}{\multirow{2}{*}{Acce-Depth}} & Acce  & \multicolumn{1}{c}{55.28} & \multicolumn{1}{c}{\textbf{72.68}} & 65.34 \\  
\multicolumn{1}{|c|}{}                            & Depth & \multicolumn{1}{c}{58.73} & \multicolumn{1}{c}{\textbf{75.22}} & 65.35 \\  \cline{1-2}
\multicolumn{1}{|c|}{\multirow{2}{*}{RGB-Depth}}  & RGB   & \multicolumn{1}{c}{72.96} & \multicolumn{1}{c}{\textbf{77.87}} & 63.26 \\  
\multicolumn{1}{|c|}{}                            & Depth & \multicolumn{1}{c}{58.21} & \multicolumn{1}{c}{\textbf{70.57}} & 67.86 \\ \hline
\end{tabular}
\label{Rstep_multimodal}
\end{center}
\end{table}

% \usepackage{multirow}
\begin{table}[t]
\caption{The comparison of average accuracy of FedMEKT under different EKT Steps on UR Fall Detection Dataset~\citep{kwolek2014human} with mixed clients. }
\begin{center}
\begin{tabular}{|cc|ccc|}
\hline
\multicolumn{2}{|c|}{\multirow{2}{*}{Modality}}                   & \multicolumn{3}{c|}{\# of EKT Steps R}                                   \\ \cline{3-5} 
\multicolumn{2}{|c|}{}                                    & \multicolumn{1}{c}{1}     & \multicolumn{1}{c}{2}              & 3     \\ \hline
\multicolumn{1}{|c|}{\multirow{2}{*}{Acce-RGB}}   & Acce  & \multicolumn{1}{c}{58.21} & \multicolumn{1}{c}{\textbf{73.92}} & 72.22 \\  
\multicolumn{1}{|c|}{}                            & RGB   & \multicolumn{1}{c}{58.23} & \multicolumn{1}{c}{\textbf{68.18}} & 59.86 \\ \cline{1-2}
\multicolumn{1}{|c|}{\multirow{2}{*}{Acce-Depth}} & Acce  & \multicolumn{1}{c}{49.00} & \multicolumn{1}{c}{\textbf{73.70}} & 56.67 \\  
\multicolumn{1}{|c|}{}                            & Depth & \multicolumn{1}{c}{56.97} & \multicolumn{1}{c}{\textbf{71.97}} & 64.64 \\  \cline{1-2}
\multicolumn{1}{|c|}{\multirow{2}{*}{RGB-Depth}}  & RGB   & \multicolumn{1}{c}{71.88} & \multicolumn{1}{c}{\textbf{80.13}} & 63.92 \\  
\multicolumn{1}{|c|}{}                            & Depth & \multicolumn{1}{c}{58.24} & \multicolumn{1}{c}{\textbf{74.10}} & 60.76 \\ \hline
\end{tabular}
\label{Rstep_mixed}
\end{center}
\end{table}

\begin{table}[t]
\caption{The comparison of average accuracy of FedMEKT under different local epochs on UR Fall Detection Dataset~\citep{kwolek2014human} with multimodal clients. }
\begin{center}
\begin{tabular}{|cc|ccc|}
\hline
\multicolumn{2}{|c|}{\multirow{2}{*}{Modality}}                   & \multicolumn{3}{c|}{\# of Local Epochs N}                                   \\ \cline{3-5} 
\multicolumn{2}{|c|}{}                                    & \multicolumn{1}{c}{1}     & \multicolumn{1}{c}{2}              & 3     \\ \hline
\multicolumn{1}{|c|}{\multirow{2}{*}{Acce-RGB}}   & Acce  & \multicolumn{1}{c}{56.71} & \multicolumn{1}{c}{\textbf{70.66}} & 68.03 \\  
\multicolumn{1}{|c|}{}                            & RGB   & \multicolumn{1}{c}{58.23} & \multicolumn{1}{c}{\textbf{66.70}} & 63.93 \\ \cline{1-2}
\multicolumn{1}{|c|}{\multirow{2}{*}{Acce-Depth}} & Acce  & \multicolumn{1}{c}{58.96} & \multicolumn{1}{c}{\textbf{72.68}} & 62.26 \\  
\multicolumn{1}{|c|}{}                            & Depth & \multicolumn{1}{c}{59.66} & \multicolumn{1}{c}{\textbf{75.22}} & 67.67 \\  \cline{1-2}
\multicolumn{1}{|c|}{\multirow{2}{*}{RGB-Depth}}  & RGB   & \multicolumn{1}{c}{67.30} & \multicolumn{1}{c}{\textbf{77.87}} & 69.37 \\  
\multicolumn{1}{|c|}{}                            & Depth & \multicolumn{1}{c}{68.48} & \multicolumn{1}{c}{\textbf{70.57}} & 61.60 \\ \hline
\end{tabular}
\label{Nstep_multimodal}
\end{center}
\end{table}

\begin{table}[t]
\caption{The comparison of average accuracy of FedMEKT under different local epochs on UR Fall Detection Dataset~\citep{kwolek2014human} with mixed clients. }
\begin{center}
\begin{tabular}{|cc|ccc|}
\hline
\multicolumn{2}{|c|}{\multirow{2}{*}{Modality}}                   & \multicolumn{3}{c|}{\# of Local Epochs N}                                   \\ \cline{3-5} 
\multicolumn{2}{|c|}{}                                    & \multicolumn{1}{c}{1}     & \multicolumn{1}{c}{2}              & 3     \\ \hline
\multicolumn{1}{|c|}{\multirow{2}{*}{Acce-RGB}}   & Acce  & \multicolumn{1}{c}{67.76} & \multicolumn{1}{c}{\textbf{73.92}} & 58.25 \\  
\multicolumn{1}{|c|}{}                            & RGB   & \multicolumn{1}{c}{60.00} & \multicolumn{1}{c}{\textbf{68.18}} & 65.14 \\ \cline{1-2}
\multicolumn{1}{|c|}{\multirow{2}{*}{Acce-Depth}} & Acce  & \multicolumn{1}{c}{57.60} & \multicolumn{1}{c}{\textbf{73.70}} & 61.42 \\  
\multicolumn{1}{|c|}{}                            & Depth & \multicolumn{1}{c}{63.65} & \multicolumn{1}{c}{71.97} & \textbf{74.58} \\  \cline{1-2}
\multicolumn{1}{|c|}{\multirow{2}{*}{RGB-Depth}}  & RGB   & \multicolumn{1}{c}{65.94} & \multicolumn{1}{c}{\textbf{80.13}} & 72.64 \\  
\multicolumn{1}{|c|}{}                            & Depth & \multicolumn{1}{c}{58.63} & \multicolumn{1}{c}{\textbf{74.10}} & 66.81 \\ \hline
\end{tabular}
\label{Nstep_mixed}
\end{center}
\end{table}

\begin{table}[t]
\caption{Results on ablation studies with multimodal clients.}
\begin{center}
\resizebox{9cm}{!}{%
\begin{tabular}{|c|cc|cc|cc|}
\hline
\multirow{2}{*}{Methods}                                           & \multicolumn{2}{c|}{Acce-RGB}                        & \multicolumn{2}{c|}{Acce-Depth}                      & \multicolumn{2}{c|}{RGB-Depth}                       \\ \cline{2-7} 
                                                                   & \multicolumn{1}{c|}{Acce}           & RGB            & \multicolumn{1}{c|}{Acce}           & Depth          & \multicolumn{1}{c|}{RGB}            & Depth          \\ \hline
FedMEKT                                                            & \multicolumn{1}{c|}{\textbf{70.66}} & \textbf{66.70} & \multicolumn{1}{c|}{\textbf{72.68}} & \textbf{75.22} & \multicolumn{1}{c|}{\textbf{77.87}} & \textbf{70.57} \\ \hline
\begin{tabular}[c]{@{}c@{}}FedMEKT\\ (w/o local EKD)\end{tabular}  & \multicolumn{1}{c|}{67.32}          & 58.21          & \multicolumn{1}{c|}{68.02}          & 74.25          & \multicolumn{1}{c|}{63.78}          & 58.21          \\ \hline
\begin{tabular}[c]{@{}c@{}}FedMEKT\\ (w/o global EKD)\end{tabular} & \multicolumn{1}{c|}{69.36}          & 58.20          & \multicolumn{1}{c|}{72.30}          & 66.06          & \multicolumn{1}{c|}{45.63}          & 58.46          \\ \hline
\end{tabular}
}%
\label{ablation_multimodal}
\end{center}
\end{table}
% \usepackage{multirow}
\begin{table}[!]
\caption{Results on ablation studies with mixed clients.}
\begin{center}
\resizebox{9cm}{!}{%
\begin{tabular}{|c|cc|cc|cc|}
\hline
\multirow{2}{*}{Methods}                                           & \multicolumn{2}{c|}{Acce-RGB}                        & \multicolumn{2}{c|}{Acce-Depth}                      & \multicolumn{2}{c|}{RGB-Depth}                       \\ \cline{2-7} 
                                                                   & \multicolumn{1}{c|}{Acce}           & RGB            & \multicolumn{1}{c|}{Acce}           & Depth          & \multicolumn{1}{c|}{RGB}            & Depth          \\ \hline
FedMEKT                                                            & \multicolumn{1}{c|}{\textbf{73.92}} & \textbf{68.18} & \multicolumn{1}{c|}{\textbf{73.70}} & \textbf{71.97} & \multicolumn{1}{c|}{\textbf{80.13}} & \textbf{74.10} \\ \hline
\begin{tabular}[c]{@{}c@{}}FedMEKT\\ (w/o local EKD)\end{tabular}  & \multicolumn{1}{c|}{70.38}          & 59.08          & \multicolumn{1}{c|}{66.48}          & 67.05          & \multicolumn{1}{c|}{71.12}          & 60.17          \\ \hline
\begin{tabular}[c]{@{}c@{}}FedMEKT\\ (w/o global EKD)\end{tabular} & \multicolumn{1}{c|}{61.59}          & 60.36          & \multicolumn{1}{c|}{73.61}          & 67.99          & \multicolumn{1}{c|}{45.70}          & 58.22          \\ \hline
\end{tabular}
}
\label{ablation_mixed}
\end{center}
\end{table}

\subsubsection{Efficiency Evaluation}
Inspired by previous work~\citep{zhang2021parameterized}, we evaluate the communication cost by recording three criteria: proxy data and knowledge size for distillation-based methods, and model parameters for model parameter-based methods. As shown in Fig.~\ref{cost}, our proposed distillation-based FedMEKT with the knowledge transfer scheme can save more communication cost than other parameter-based methods.

\subsubsection{Effect of Proxy Dataset}
In this experiment, we evaluate the performance of our proposed method which utilizes proxy data for exchanging knowledge between the server and clients. We assess the impact of varying the size of the proxy dataset, $D_r$, on the FedMEKT-C algorithm with different client settings. Tables~\ref{multimodal_proxy_data} and~\ref{mixed_proxy_data} demonstrate the effect of the proxy data size on the algorithm's performance. The results indicate that increasing the size of the proxy data $D_r$ enhances the performance in most of the scenarios. In our experiment on the UR Fall Detection Dataset~\citep{kwolek2014human}, we set the default size of the proxy dataset to $1000$ samples, which corresponds to $100\%$ of $D_r$ and $1/10$ of the total data.
\subsubsection{Effect of Hyperparameters}
\paragraph{Effect of Knowledge Transfer Step R} In this experiment, we compare the performance of FedMEKT on the UR Fall Detection dataset~\citep{kwolek2014human} under different settings of embedding knowledge transfer (EKT) steps. The number of local epochs is set to $2$. As shown in Tables~\ref{Rstep_multimodal} and~\ref{Rstep_mixed}, we find that the value $R=2$ consistently yields the highest performance across all scenarios, outperforming other values.

\paragraph{Effect of Local Epoch N} Similarly, we also investigate the behavior of the local epoch $N$ on the UR Fall Detection dataset~\citep{kwolek2014human}, as shown in Tables~\ref{Nstep_multimodal} and~\ref{Nstep_mixed}, with the number of knowledge transfer steps fixed at $2$. Our findings indicate that achieving optimal performance requires a moderate number of local epochs, as a large value of $N$ does not always result in better performance.

\paragraph{Other Hyperparameter Tuning} We tune the hyperparameters $\gamma$ and $\beta$ for FedMEKT, selecting from the range of \{0.01, 0.1\} for each modality combination, and report the best result. For MM-FedProx and MM-MOON, we adopt the values from the original works~\citep{li2020federated,li2021model} and tune the hyperparameter $\mu$ for the proximal term and the contrastive loss. We use the Adam optimizer with a learning rate in the range between \{0.001, 0.01\} for all approaches and report the best result.

\subsubsection{Ablation Studies}
In this experiment, we investigate the impact of EKD regularizer on both the server and client sides. We conduct experiments on UR Fall Detection~\citep{kwolek2014human} under different scenarios, as shown in Table~\ref{ablation_multimodal} and \ref{ablation_mixed}, by removing the EKD regularizer on local and global sides, respectively. Our results demonstrate that FedMEKT can significantly benefit from the EKD regularizer on both sides of the framework. In all use cases, we observe the improved performance with the EKD regularizer present on both sides.

% \section{Discussion}
% In this section, we explore the strengths and limitations of our method in comparison to MM-FedAvg and other related works. Firstly, FedMEKT offers several advantages, including communication cost savings by avoiding parameter aggregation and achieving higher performance in multimodal FL. By utilizing joint embedding knowledge exchange instead of parameter exchange, FedMEKT significantly reduces the communication cost required between the server and clients. Furthermore, FedMEKT demonstrates superior performance compared to other baselines. The joint embedding knowledge transfer mechanism allows the global model to capture the distinctive features of multimodal data from all local models, leading to enhanced generalization capabilities and improved accuracy in downstream tasks. However, it is essential to acknowledge that FedMEKT also has limitations, such as the requirement of a proxy dataset, which can be cumbersome for tasks involving large amounts of data. In scenarios where the sample number is extremely large, and the task complexity is high, both FedMEKT and other baseline methods would still incur a high communication cost. Future research could explore strategies to address this limitation and optimize communication efficiency in multimodal FL.

\section{Conclusion}
In this work, we proposed a novel multimodal federated learning framework under the semi-supervised setting by developing the joint embedding knowledge transfer scheme. FedMEKT offers higher performance in multimodal FL, while saving communication cost by avoiding parameter aggregation. By utilizing joint embedding knowledge transfer instead of parameter exchanging, FedMEKT significantly reduces the communication cost required between the server and clients and prevents the reverse engineering. Furthermore, FedMEKT demonstrates superior performance compared to other baselines. The joint embedding knowledge transfer mechanism allows the global model to capture the distinctive features of multimodal data from all local models, leading to enhanced generalization capabilities and improved accuracy in downstream tasks. Through extensive simulations, our proposed FedMEKT obtains a more stable and better performance in downstream tasks than other baselines without exchanging model parameters. This results in saving communication cost, particularly in large models with millions of parameters, enhancing privacy protection. Future research will extend this work to include other tasks and additional modalities, enabling the deployment of federated learning in personalized applications.
\bibliographystyle{elsarticle-harv} 
\bibliography{egbib}
\appendix
\section{Formulation for Algorithms Under Multimodal FL Framework}
Inspired by other state-of-the-art FL methods such as FedProx~\citep{li2020federated} and MOON~\citep{li2021model}, we reproduce several algorithms under multimodal federated learning framework. For each algorithm, we present its mathematical formulation.
\subsection{MM-FedAvg}
 We denote $n_A=\sum_{k\in{m_{A}, m_{AB}}}{n_k}$,  $n_B=\sum_{k\in{m_{B}, m_{AB}}}{n_k}$ as the total number of data samples for the modality $A$ and $B$, respectively, where $m$ denotes the set of clients in each modality (i.e., client with modality $m_{AB}$ holds both data from modality $A$ and $B$), and $n_k$ is the number of data samples for client $k$. The learning objective functions for global autoencoders A and B of MM-FedAvg~\citep{zhao2022multimodal} can be defined as:
\begin{equation}
\begin{split}
     \min_{f_{A},g_{A}}{\mathcal{L}}_{s}(f_{A},g_{A})= \sum_{k\in{m_{A}}}{\frac{n_{k}}{n_{A}}}&{\mathcal{L}}_{k}(f_{A},g_{A}) \\
     &+\alpha \, \sum_{k\in{m_{AB}}}{\frac{n_{k}}{n_{A}}}{\mathcal{L}}_{k}(f_{A},g_{A}),  
    % \text{where}  {\ell}^{}_{k}(f_{A},g_{A})=\min_{f_{A},g_{A}}{\ell}^{}_{A}(x_A,x_A') +{\ell}_{A}(x_B, x_B')
\end{split}
\end{equation}

\begin{equation}
\begin{split}
       \min_{f_{B},g_{B}}{\mathcal{L}}^{}_{s}(f_{B},g_{B})= \sum_{k\in{m_{B}}}{\frac{n_{k}}{n_{B}}}&{\mathcal{L}}_{k}(f_{B},g_{B}) \\
       &+\alpha \, \sum_{k\in{m_{AB}}}{\frac{n_{k}}{n_{B}}}{\mathcal{L}}_{k}(f_{B},g_{B}),  
    % \text{where}  {\ell}^{}_{k}(f_{A},g_{A})=\min_{f_{A},g_{A}}{\ell}^{}_{A}(x_A,x_A') +{\ell}_{A}(x_B, x_B')
\end{split}
\end{equation}
where ${\mathcal{L}}^{}_{k}(f_{A},g_{A})$ and ${\mathcal{L}}^{}_{k}(f_{B},g_{B})$ are loss functions of split autoencoder for each modality $A$ and $B$. Specifically, loss functions at client $k$ for MM-FedAvg can be defined as:
\begin{equation}
    {\mathcal{L}}^{}_{k}(f_{A},g_{A})= \min_{f_{A},g_{A}}{\ell}^{}_{A}(x^A,x^{A'}) +{\ell}_{B}(x^B, x^{B'}), 
\end{equation}

\begin{equation}
    {\mathcal{L}}^{}_{k}(f_{B},g_{B})= \min_{f_{B},g_{B}}{\ell}^{}_{A}(x^A,x^{A'}) +{\ell}_{B}(x^B, x^{B'}),
\end{equation}
where $\ell_A$ and $\ell_B$ are reconstruction losses \citep{zhao2022multimodal} (e.g., MSE loss).
\subsection{MM-FedProx}
Similar to MM-FedAvg~\citep{zhao2022multimodal}, the learning objective functions for global autoencoders A and B can be defined as:
\begin{equation}
\begin{split}
     \min_{f_{A},g_{A}}{\mathcal{L}}_{s}(f_{A},g_{A})= \sum_{k\in{m_{A}}}{\frac{n_{k}}{n_{A}}}&{\mathcal{L}}_{k}(f_{A},g_{A}) \\
     &+\alpha \, \sum_{k\in{m_{AB}}}{\frac{n_{k}}{n_{A}}}{\mathcal{L}}_{k}(f_{A},g_{A}),  
    % \text{where}  {\ell}^{}_{k}(f_{A},g_{A})=\min_{f_{A},g_{A}}{\ell}^{}_{A}(x_A,x_A') +{\ell}_{A}(x_B, x_B')
\end{split}
\end{equation}

\begin{equation}
\begin{split}
       \min_{f_{B},g_{B}}{\mathcal{L}}^{}_{s}(f_{B},g_{B})= \sum_{k\in{m_{B}}}{\frac{n_{k}}{n_{B}}}&{\mathcal{L}}_{k}(f_{B},g_{B}) \\
       &+\alpha \, \sum_{k\in{m_{AB}}}{\frac{n_{k}}{n_{B}}}{\mathcal{L}}_{k}(f_{B},g_{B}),  
    % \text{where}  {\ell}^{}_{k}(f_{A},g_{A})=\min_{f_{A},g_{A}}{\ell}^{}_{A}(x_A,x_A') +{\ell}_{A}(x_B, x_B')
\end{split}
\end{equation}
where ${\mathcal{L}}^{}_{k}(f_{A},g_{A})$ and ${\mathcal{L}}^{}_{k}(f_{B},g_{B})$ are loss functions of split autoencoder for each modality $A$ and $B$. Specifically, loss functions at client $k$ for MM-FedProx can be defined as:
\begin{equation}
\begin{split}
    {\mathcal{L}}^{}_{k}(f_{A},g_{A})= \min_{f_{A},g_{A}}{\ell}^{}_{A}(x^A,x^{A'}) +{\ell}_{B}(x^B, x^{B'})
    +\mu\ell_{proxA}, 
\end{split}
\end{equation}

\begin{equation}
\begin{split}
    {\mathcal{L}}^{}_{k}(f_{B},g_{B})= \min_{f_{B},g_{B}}{\ell}^{}_{A}(x^A,x^{A'}) +{\ell}_{B}(x^B, x^{B'}) 
    +\mu\ell_{proxB},
\end{split}
\end{equation}
where $\ell_A$ and $\ell_B$ are reconstruction losses \citep{zhao2022multimodal} (e.g., MSE loss), $\ell_{proxA}=\left \| w_s^A - w_k^A \right \|^{2}$, $\ell_{proxB}=\left \| w_s^B - w_k^B \right \|^{2}$ while $x^{A'}$ and $x^{B'}$ denote the reconstructed outputs of two modalities A and B, respectively.
\subsection{MM-MOON}
Following the MM-FedAvg~\citep{zhao2022multimodal} mechanism, MM-MOON assigned higher weights (set to 100) to multimodal clients in model aggreagation. The learning objective functions for global autoencoders A and B can be defined as:
\begin{equation}
\begin{split}
     \min_{f_{A},g_{A}}{\mathcal{L}}_{s}(f_{A},g_{A})= \sum_{k\in{m_{A}}}{\frac{n_{k}}{n_{A}}}&{\mathcal{L}}_{k}(f_{A},g_{A}) \\
     &+\alpha \, \sum_{k\in{m_{AB}}}{\frac{n_{k}}{n_{A}}}{\mathcal{L}}_{k}(f_{A},g_{A}),  
    % \text{where}  {\ell}^{}_{k}(f_{A},g_{A})=\min_{f_{A},g_{A}}{\ell}^{}_{A}(x_A,x_A') +{\ell}_{A}(x_B, x_B')
\end{split}
\end{equation}

\begin{equation}
\begin{split}
       \min_{f_{B},g_{B}}{\mathcal{L}}^{}_{s}(f_{B},g_{B})= \sum_{k\in{m_{B}}}{\frac{n_{k}}{n_{B}}}&{\mathcal{L}}_{k}(f_{B},g_{B}) \\
       &+\alpha \, \sum_{k\in{m_{AB}}}{\frac{n_{k}}{n_{B}}}{\mathcal{L}}_{k}(f_{B},g_{B}),  
    % \text{where}  {\ell}^{}_{k}(f_{A},g_{A})=\min_{f_{A},g_{A}}{\ell}^{}_{A}(x_A,x_A') +{\ell}_{A}(x_B, x_B')
\end{split}
\end{equation}
where ${\mathcal{L}}^{}_{k}(f_{A},g_{A})$ and ${\mathcal{L}}^{}_{k}(f_{B},g_{B})$ are loss functions of split autoencoder for each modality $A$ and $B$. Specifically, loss functions at client $k$ for MM-MOON can be defined as:
\begin{equation}
\begin{split}
    &{\mathcal{L}}^{}_{k}(f_{A},g_{A})= \min_{f_{A},g_{A}}{\ell}^{}_{A}(x^A,x^{A'}) +{\ell}_{B}(x^B, x^{B'})
    +\mu\ell_{conA},
\end{split}
\end{equation}

\begin{equation}
\begin{split}
    {\mathcal{L}}^{}_{k}(f_{B},g_{B})= \min_{f_{B},g_{B}}{\ell}^{}_{A}(x^A,x^{A'}) +{\ell}_{B}(x^B, x^{B'}) 
    +\mu\ell_{conB},
\end{split}
\end{equation}
where $\ell_A$ and $\ell_B$ are reconstruction losses \citep{zhao2022multimodal} (e.g., MSE loss), $\ell_{conA}$ and $\ell_{conB}$ are the model-contrastive loss for modality A and B, respectively. The model-contrastive loss for two modalities are denoted as:
{
\small
\begin{equation}
% \begin{aligned}
    \ell_{conA}  = -\log{\frac{\exp(\frac{\textrm{sim}(z_A, z_{globA})}{\tau})}{\exp(\frac{\textrm{sim}(z_A, z_{globA})}{\tau}) + \sum_{i=1}^k\exp(\frac{\textrm{sim}(z_A, z_{prevA}^i)}{\tau})}}
% \end{aligned}
\end{equation}
}

{
\small
\begin{equation}
% \begin{aligned}
    \ell_{conB}  = -\log{\frac{\exp(\frac{\textrm{sim}(z_B, z_{globB})}{\tau})}{\exp(\frac{\textrm{sim}(z_B, z_{globB})}{\tau}) + \sum_{i=1}^k\exp(\frac{\textrm{sim}(z_B, z_{prevB}^i)}{\tau})}}
% \end{aligned}
\end{equation}
}
where $\tau$ denotes a temperature parameter. We follows the original paper MOON~\citep{li2021model} for setting $\tau$.
\subsection{FedMEKT-S}
Contrast to our primary variant FedMEKT-C, FedMEKT-S adopts the split embedding knowledge from each modality and updates the autoencoders of two modalities A and B asynchronously. The global loss functions for global autoencoders A and B can be defined as: 
\begin{equation}
\begin{split}
       {\mathcal{L}}^s_{A}= {\ell}_{A}(x^A_r,g_A(h_A)|D_r) & +{\ell}_{B}(x^B_r,g_B(h_A)|D_r) \\
       &+\beta \, {\ell}_{EKD}\bigg(e^{Ah}_{s},\sum_{k\in{K}}{\frac{1}{K}}e^{Ah}_{k}|D_r\bigg),
       \label{eq:globalA-mmEKT}
\end{split}
\end{equation}

\begin{equation}
\begin{split}
    {\mathcal{L}}^s_{B}= {\ell}_{A}(x^A_r,g_A(h_B)|D_r)& +{\ell}_{B}(x^B_r,g_B(h_B)|D_r) \\
    &+\beta \, {\ell}_{EKD}\bigg(e^{Bh}_{s},\sum_{k\in{K}}{\frac{1}{K}}e^{Bh}_{k}|D_r\bigg),
        \label{eq:globalB-mmEKT}
\end{split}
\end{equation}
where $\ell_A$ and $\ell_B$ are reconstruction~\citep{zhao2022multimodal} (e.g., MSE loss), $\ell_EKD$ is a embedding knowledge transfer regularizer. Different from FedMEKT-C, the variant FedMEKT-S updates the autoencoders of each modality sequentially. The local loss functions for local autoencoders are denoted as follow:
\begin{equation}
\begin{split}
  {\mathcal{L}}^{k}_{A}= {\ell}_{A}(x^A_k,g_A(h_A)|D_k) &+{\ell}_{B}(x^B_k,g_B(h_A)|D_k) \\
  &+\gamma \, {\ell}_{EKD}(e^{Ah}_{k},e^{Ah}_{s}|D_r),
    \label{eq:ondeviceA-mmEKT}
\end{split}
\end{equation}
\begin{equation}
\begin{split}
  {\mathcal{L}}^{k}_{B}= {\ell}_{A}(x^A_k,g_A(h_B)|D_k) &+{\ell}_{B}(x^B_k,g_B(h_B)|D_k) \\
  &+\gamma \, {\ell}_{EKD}(e^{Bh}_{k},e^{Bh}_{s}|D_r),
     \label{eq:ondeviceB-mmEKT}
\end{split}
\end{equation}
Similar with global model, we apply the asynchronous update for the local autoencoder of two modalities A and B.






% \end{thebibliography}
\end{document}